# Comparative Study of Language Models on Cross-Domain Data with Model Agnostic Explainability

Mayank Chhipa
Fidelity Investments
Bangalore, India
mayanksamariya99@gmail.com

Hrushikesh Mahesh Vazurkar
Fidelity Investments
Bangalore, India
hrushi1999@gmail.com

Abhijeet Kumar
Emerging Tech Team
Fidelity Investments
Bangalore, India
Abhijeet.Kumar@fmr.com

Mridul Mishra
Emerging Tech Team
Fidelity Investments
Bangalore, India
Mridul.Mishra@fmr.com

*Abstract*— With the recent influx of bidirectional contextualized transformer language models in the NLP, it becomes a necessity to have a systematic comparative study of these models on variety of datasets. Also, the performance of these language models has not been explored on non-GLUE datasets. The study presented in paper compares the state-of-the-art language models - BERT, ELECTRA and its derivatives which include RoBERTa, ALBERT and DistilBERT. We conducted experiments by finetuning these models for cross domain and disparate data and penned an in-depth analysis of model's performances. Moreover, an explainability of language models coherent with pretraining is presented which verifies the context capturing capabilities of these models through a model agnostic approach. The experimental results establish new state-of-the-art for Yelp 2013 rating classification task and Financial Phrasebank sentiment detection task with 69% accuracy and 88.2% accuracy respectively. Finally, the study conferred here can greatly assist industry researchers in choosing the language model effectively in terms of performance or compute efficiency.

## I. INTRODUCTION

With the advent of sequence models in deep learning, NLP tasks had improved the prediction metrics with the usage of Recurrent neural networks (RNN). A variation of RNN architecture, Long Short-Term Memory networks (LSTM) were implemented to deal with long term dependencies by introducing a memory cell into the network. This firmly established the RNN, LSTM and Gated recurrent neural networks (GRU) as the state-of-the-art approaches in sequence modeling [1]. Much recently with the advent of transformers, the recurrent layers in encoder-decoder architectures were replaced with multi headed self-attention [2]. These language models were based on transfer learning and restricted the power of pre-trained representations, especially for fine tuning approaches [3]. The major limitation of standard language models was their unidirectional nature, i.e., they still processed context only in one direction (sequence). In recent years, there has been great research attention in development of bidirectional contextualized transformer language representation models. Unlike the conventional unidirectional language representation models, these models pre-train deep bidirectional representations from unlabeled text and fuse both left and right context in all layers [3].

BERT improved the finetuning based approaches by alleviating the unidirectionality constraint [3]. RoBERTa presented a replication study of BERT with careful evaluation of the pre-training hyperparameters and training set size. It was found that BERT was significantly undertrained, and performance was significantly enhanced with the improved training procedure [4]. As transfer learning from large scale pre-trained models became more prevalent in NLP, DistilBERT, a distilled version of BERT was introduced. DistilBERT enabled operating these large models under constrained computational resources [5]. Increasing model size to improve performance becomes harder with GPU/TPU memory limitations and results in larger training times. ALBERT addressed these problems by presenting two parameter sharing techniques to lower memory consumption and increase training speed [6]. The most recent language model ELECTRA provides enhanced compute efficiency and better performance on downstream NLP tasks [7]. In this paper, we finetune these language models on disparate data and presented a detailed study. The chosen datasets have different properties in order to establish the fact that these models are robust to any kind of data. For instance, Financial Phrasebank dataset is highly domain dependent and context sensitive whereas Yelp dataset is noisy with skewed ratings distribution [8][9].

With success of language models in all the experimented datasets, we also performed experiments to explain the language models and found intriguing results. Since these models work like a black box, in order to understand the kind of context learnt and gain deeper insights about the learning of models, we depict an approach for the interpretability of language models using the financial phrasebank dataset. Our main work is summarized as follows:

- Comprehensive and comparative analysis of the performance of the transformer language models on non-GLUE datasets [10].

- We achieved new state-of-the-art results on the Yelp2013 and Financial Phrasebank datasets with accuracy of 69% and 88.2% respectively.

- Describes a model independent interpretability approach to explain the context capturing potential of black box language models.

This work was supported by Asset Management Group, Fidelity Investment. Disclaimer: The views or opinions expressed in this paper are solely those of the author and do not necessarily represent those of Fidelity Investments. This research does not reflect in any way procedures, processes or policies of operations within Fidelity Investments. (Mayank Chhipa and Hrushikesh Mahesh Vazurkar are co-first authors.)

## II. RELATED WORK

There has been extensive research work in the field of BERT-based (derived) language models since its release in 2018. Since these models are relatively new and have not been tested much on non-GLUE datasets, several experiments employing these language models had been performed on SemEval 8, Financial Phrasebank and Yelp 2013 datasets for comparison with previous state-of-the-art results.

Srikumar employed hierarchical sentiment classifier using performance indicators for financial sentiment prediction and achieved an accuracy of over 80% [11]. Dogu Arci et al. further pre-trained BERT on additional financial domain data and finetuned on Financial Phrasebank dataset to get state-of-the-art accuracy of 86% [8]. Yelp 2013 is a dataset consisting of user reviews and corresponding rating. Jihyeok and Ampalyo use basis vectors to effectively incorporate categorical metadata on various parts of a neural-based model and achieves 67.1% accuracy on Yelp 2013 [12][13]. Amplayo further proposed BiLSTM+CHIM to represent attributes as chunk-wise importance weight matrices and consider four locations in the model (i.e., embedding, encoding, attention, classifier) to inject attributes achieving the current published state-of-the-art accuracy of 67.8 [14].

Relation classification is a crucial ingredient in numerous information extraction systems seeking to mine structured facts from text. Semeval 2010 Task 8 dataset focused on semantic relations between pairs of nominals [15]. Livio Baldini et al. built on extensions of Harris' distributional hypothesis to relations, as well as recent advances in learning text representations (specifically, BERT), to build task agnostic relation representations solely from entity-linked text [16]. Cheng Li et al. proposed REDN designed with new network architecture along with a special loss function designed to serve as a downstream model of Language Models for supervised relation extraction, achieving F1 score of 91.0 [17].

There are currently various tools for explainability of black box. LIVE and LIME packages use surrogate models (the so-called white box models) that approximate local structure of the complex black box model [18][19]. Shapely, another unified framework based on results from game theory uses improved DeepLift algorithm to explain TensorFlow based models [19][20][21]. breakDown, R library uses a greedy approach in which only a single series of nested conditionings is considered [18]. Even though there had been a surge in research for AI explainability, there is dearth of implemented frameworks for supporting explainability of transformer-based language models like BERT. There had been recent efforts made to study the self-attention mechanism (transformer layer) of language models but it has not been evidently conclusive in terms of important features [22][23][24].

The outline of the rest of paper is as follows: The next section briefly describes 5 much recent language models accounted and our experimental setup with 5 different datasets. Section 4 shows the results, comparative study and its analysis. In section 5, we demonstrate an explainability route taken to visualize the important features (context) from the language model.

## III. EXPERIMENTS

### A. Language Models

- **BERT**: Pre-trained on Wikipedia (2500M words) and Books Corpus (800M words) for two tasks i.e. masked language modeling (MLM) and next sentence prediction (NSP) tasks. BERT-base model with 110M parameters was finetuned for our experiments [3].

- **ELECTRA**: The pre-training data and number of parameters of ELECTRA-base is same as BERT-base. Replace token detection task is employed instead of MLM which makes it more compute efficient and more accurate in terms of performance [7].

- **DistilBERT**: It leverages knowledge distillation in pre-training phase to reduce the size of BERT-base by 40%, while retaining 97% of its language understanding capabilities and being 60% faster with 66M number of parameters [5].

- **ALBERT**: A lite model based on BERT architecture and same pre-training corpus. ALBERT-base uses 12M parameters only as it achieves significant parameter reduction due to factorized embedding parameterization and cross layer parameter sharing. ALBERT introduced a self-supervised loss for sentence-order prediction (SOP) to address the ineffectiveness of NSP task in BERT [6].

- **RoBERTa**: Pre-trained on CC News corpus apart from the BERT data with 125M parameters. It compromised on the NSP objective and pre-trained with larger mini batches and learning rate to improve upon MLM [4].

An optimized SVM classifier was also included in the experiments to set the baseline for comparison. The experiments were conducted using Google's Tensorflow implementation of BERT and ELECTRA and HuggingFace PyTorch Transformers library for DistilBERT, ALBERT and RoBERTa [25][26][27].

### B. Model Hyperparameters

To choose the best hyperparameters for effective finetuning, grid search approach was used to search for parameters which could give the best results, unleashing the full potential of models.

TABLE I. MODEL HYPERPARAMETERS

| Dataset | Hyperparameters | | |
|---|---|---|---|
| | *Maximum Sequence Length* | *Batch size* | *Epochs* |
| BBC news | 256 | 16 | 4 |
| BBc sports | 256 | 16 | 4 |
| Financial PhraseBank | 64 | 16 | 4 |
| Yelp 2013 | 256 | 16 | 3 |
| SemEval Task 8 | 64 | 64 | 5/6 |

Exploratory data analysis resulted in adoption of correct maximum sequence length to minimize information loss. A learning rate of 5e-5 have been used with a warmup proportion of 0.1 for all experiments. Table. 1 present the optimal hyperparameters chosen for all the models in consideration across different datasets.

*C. Datasets*

These datasets include BBC News, BBC Sports, Financial Phrasebank, Yelp 2013 and SemEval Task 8 dataset. These datasets are disparate in nature and have different levels of complexity and context sensitivity. For example, BBC News and BBC sports are context-insensitive (independent) datasets whereas financial phrasebank is a highly context sensitive dataset related to finance domain [28][29]. Apart from these, Yelp is a noisy reviews dataset where user and product information of each review is available. We refer to it as noisy as the reviews contain noise in form of user sensitive review text [13]. For example, "good" might mean a 5 rating for a certain user but not for the other. SemEval is a complex dataset with underlying relationship between marked related entities in the sentences [15]. The details of each dataset are as follows:

*1) BBC News:* The dataset consists of 2225 documents from the BBC news website corresponding to stories in five tropical areas from 2004-2005. There are 5 categories of news articles from business (510), entertainment (386), politics (417), sport (511) and tech (401) [29].

*2) BBC Sports:* The dataset consists of 737 documents from the BBC Sport website corresponding to sports news articles in five tropical areas from 2004-2005. These areas include athletics (101), cricket (124), football (265), rugby(147) and tennis(100) [29].

*3) Financial PhraseBank:* This financial domain dataset consists of 4845 english sentences randomly selected from financial news found on LexisNexis database [28].

*4) Yelp 2013*: Yelp dataset contains reviews of 1633 products or business given by 1631 users. The review dataset has a review rating scale from 1-5. [13].

*5) SemEval Task 8*: There are nine types of annotated relations between entities along with demarcation of entities [12]. There is also a relation type "Other" (artificial class) which indicates that the relation expressed in the sentence is not among the nine types. In this dataset, each of the relation types, directionality information is also present which implies that relationship between entities can be from e1 to e2 or vice versa. For example, Entity-Origin (e1, e2) and Entity-Origin (e2, e1) can be considered two distinct relations, so the total number of relationship classes is 19 [15].

*D. Evaluation Metrics*

We split each corpus into train, validation and test sets. For initial three datasets mentioned in previous section, the test data contains 20% of the total documents and then the rest of data is split into 80% for training and 20% for validation. Yelp 2013 dataset consists of a training set of 62522 reviews, validation set of 7773 reviews and a test set with 8671 reviews. SemEval consists of a training set of 8,000 examples, and a test set with 2717 sentences. Evaluation metrics used were accuracy, macro-averaged precision, recall and F1-Score on the test data. In order to remove any bias of model finetuning, these metrics were an average score over 3 model iterations for each of the datasets.

IV. RESULTS AND ANALYSIS

The results of various experiments performed are presented in this section. Table. II and Table. III shows the evaluation metrics for all the language models along with optimized SVC model.

TABLE II. COMPARISON: EVALUATION RESULTS

| Model | BBC News | | | |
|---|---|---|---|---|
| | *Accuracy* | *Precision* | *Recall* | *F1-Score* |
| BERT | 98.37 | 98.30 | 98.37 | 98.37 |
| ELECTRA | 98.3 | 98.2 | 98.4 | 98.3 |
| ALBERT | 95.5 | 95.2 | 94.9 | 94.9 |
| RoBERTa | 96.2 | 96.2 | 96.3 | 96.1 |
| DistilBERT | 96 | 96 | 95.9 | 95.8 |
| SVC | 98.3 | 98.3 | 98.3 | 98.3 |

TABLE III. COMPARISON: EVALUATION RESULTS

| Model | BBC Sports | | | |
|---|---|---|---|---|
| | *Accuracy* | *Precision* | *Recall* | *F1-Score* |
| BERT | 99.3 | 99.3 | 99.2 | 99.3 |
| ELECTRA | 99.5 | 99.5 | 99.5 | 99.5 |
| ALBERT | 98 | 98.6 | 98 | 98.3 |
| RoBERTa | 95.9 | 96.8 | 95.2 | 95.8 |
| DistilBERT | 99.3 | 99 | 99.6 | 99.3 |
| SVC | 99.5 | 99.7 | 99.5 | 99.7 |

Clearly, BBC News and BBC Sports are easy and context-insensitive datasets. Our conjecture is that SVC uses bag-of-words encoding of the sentence, whereas language models generates context-sensitive embeddings from tokenized indices through self-attention mechanism [2]. Here, self-attention mechanism may introduce minor error in trying to learn context, performing an operation which is not required. Hence, SVC performed better than recent language models on these datasets though results were very similar.

TABLE IV. COMPARISON: EVALUATION RESULTS

| Model | Financial Phrasebank | | | |
|---|---|---|---|---|
| | *Accuracy* | *Precision* | *Recall* | *F1-Score* |
| BERT | 86.16 | 84.62 | 82.66 | 83.56 |
| ELECTRA | 88.2 | 86.4 | 87 | 87 |
| ALBERT | 84 | 81.8 | 81.2 | 81.5 |
| RoBERTa | 87.5 | 86.1 | 86.4 | 86.2 |
| DistilBERT | 86.3 | 84.2 | 83.2 | 83.7 |
| SVC | 72.7 | 67.1 | 67.5 | 67.2 |

Table. IV presents evaluation metrics comparison across models for Financial Phrasebank dataset. Here, performance of SVC plummets due to context-sensitive nature of the dataset. Moreover, Electra achieves new state-of-the-art accuracy of 88.2% on this limited data outperforming all the previous works [8]. Roberta performed close to Electra.

Table. V presents the performance of all models on SemEval Task 8. Note that the presented F1-score encompasses all 19 classes (including artificial 'Other' class) unlike semval-2010 offline scorer which ignores 'Other' class while scoring. SVC performed poor as it was a hard and context sensitive task to learn the underlying relationship between entities. BERT performed best among all models. The input format of training sentence with entities markers is shown below. There is a significant reduction of approximately 7-8% in performance without entity markers [30].

*The #student# $association$ is the voice of the undergraduate student population. (Label-Collection (e1, e2))*

More exhaustive hyperparameter search for Electra or Roberta can be explored for this task as the hyperparameters chosen were based on optimizing BERT from our earlier work [30].

TABLE V. COMPARISON: EVALUATION RESULTS

| Model | Semeval 2010 Task 8 | | | |
|---|---|---|---|---|
| | *Accuracy* | *Precision* | *Recall* | *F1-Score* |
| BERT | 85.07 | 80.57 | 82.57 | 81.47 |
| ELECTRA | 84.4 | 79.5 | 82.7 | 80.8 |
| ALBERT | 79.7 | 77.4 | 74.9 | 76 |
| RoBERTa | 85 | 79.9 | 81.7 | 80.7 |
| DistilBERT | 83.5 | 79.6 | 80.1 | 79.6 |
| SVC | 48.1 | 42.8 | 47.5 | 44.6 |

Table. VI displays evaluation results on Yelp 2013 dataset which is a noisy dataset. As user ratings are not bound by any guidelines and there is a high possibility of incoherency between user review and user rating which is highly dependent on preferences of the user. To tackle this issue, user attribute injection is an intuitive method to reduce the noise and address user preferences. A method to inject user and product attributes was implemented by training a random forest classifier on concatenation of output logits (from language model 'Electra'), 'user_id' and 'product_id' as training set and provided rating labels. The motivation for taking this approach was to allow model to learn the bias for discrepancies between sentence and user rating separately without tampering with the sentiment of the sentence.

TABLE VI. COMPARISON: EVALUATION RESULTS

| Model | Yelp 2013 | | | |
|---|---|---|---|---|
| | *Accuracy* | *Precision* | *Recall* | *F1-Score* |
| BERT | 64.9 | 63.8 | 61 | 62.3 |
| ELECTRA | 68.3 | 67.5 | 64.8 | 66 |
| DistilBERT | 64 | 64 | 60 | 62 |
| Electra + RF (Stacked) | 69 | 67 | 66 | 67 |

This stacked approach achieved state-of-the-art accuracy i.e 69% for Yelp 2013 dataset outperforming previous works [14]. As a single model, Electra also attains 68.3% accuracy without any attribute injection. Due to hardware constraints, we were limited to use maximum sequence length not greater than 256 and given the fact that 25% of the reviews had greater length, considerable amount of information was lost. We believe testing language model with 512 length would improve the evaluation metrics further.

## V. EXPLAINABILITY OF LANGUAGE MODELS

The idea behind this work was two folded, first to study if the language model learns the context while finetuning and secondly to compare different language models in terms of explainability of important features.

### A. Explanbility Method: Approach

Taking motivation from LIME [19], we demonstrate an interpretability technique that assumes linear contribution of each feature (word) in the confidence score of prediction. The method can be applied to any language model.

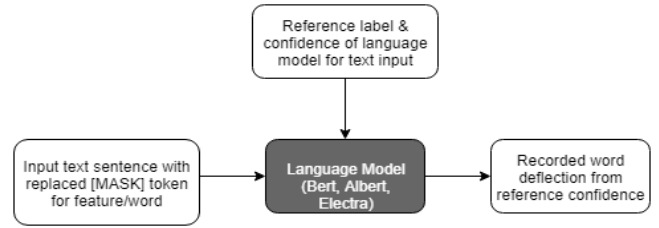

Fig. 1. Explainability of Language Models – Masking each feature in iteration and recording the variation from reference.

Each word of the input text is masked with [MASK] token iteratively and the variation in confidence score of finetuned model from reference confidence score is recorded. The reason of choosing [MASK] token was the fact that the same was used in the pre-training of language models (in BERT and variants [3][4][5][6][7]). Reference confidence score is prediction probabilities from model when no word of the text input was masked. The recorded variation can be used to order the features according to its importance. The feature(word) having highest variation from actual reference is considered most important and vice versa. It is important to note that if variation is in negative direction (lower than reference) then it is considered important feature else it is a deteriorating feature otherwise.

### B. Comparing Models: Explainable Contextual Learning

Financial PhraseBank dataset was chosen for the experiments due to our interest in finance domain and the fact that sentiment task is extremely context sensitive which makes it distinctive to look at the context learnt by the models. The interpretability can also provide insights for sentences resulting in incorrect prediction. Figure. 2 shows the words and its importance for two opposite sentiment statements (given below).

*Changes in the market situation and tougher price competition have substantially reduced demand for bread packaging manuf actured at the Kauhava plant, according to the company.*

*Of the sales price, a sales gain of some 3.1 mln euro ($4.5 mln) will be recognized in Incap 's earnings for 2007.*

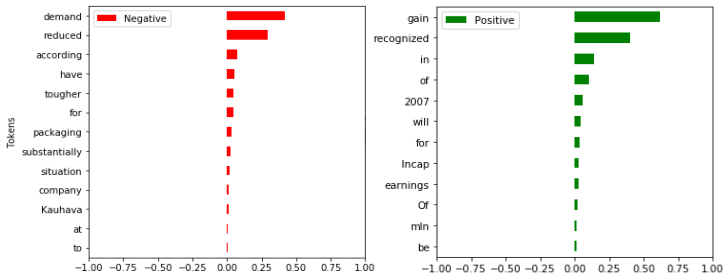

Fig. 2. Feature importances for two example sentences

We performed the explainability experiments with three language models namely Bert, Roberta and Electra. A more clear and detailed set of examples of financial statements can be seen in table provided in appendix. It depicts top 3 important features (words) sequentially in order of their importance. Brighter color denotes higher importance. Also, each row in table (refer Appendix) compares pair of sentences with similar aspect/subject (in bold) but that aspect is captured by models as important feature in opposite sentiments. The results clearly show that language models gave more attention to the contextual words responsible for polarity. It was also intriguing to look at the contexts captured across language models, we observed that Electra learns better context than Bert which justifies the higher performance on this dataset.

As the language models have ability to learn nuances (linguistic features) of language, we observed from explainability experiments that finetuning it for classification task drifts all the tokens of the sentence towards the intended label it has learnt. Moreover, the ability to learn long sequences non-linearly could be the reason for our observation in linear fashioned explainability. Finally, we observed that models also learn comparative prepositions or tenses as important feature which significantly decides polarity in financial statements e.g. *to, from, was, up, at, with, by.*

## VI. CONCULSION AND FUTURE WORK

We conducted experiments to perform a comparative study of the recent language models on disparate datasets and provide inferences about their performance. One obvious observation was the fact that nature of the dataset dictates the performance of any language model. In this paper, we presented new state-of-the-art accuracy achieved by Electra on two tasks i.e. user rating classification on Yelp 2013 and sentiment detection on Financial Phrasebank with 69% and 88.2% respectively.

Majorly there are two conclusions from the analysis. First, Electra performs best among all language models on most of the datasets. Secondly, Albert is very compute efficient and despite being less parametric (reduction in order of 10) gives comparable performance. Also, the paper introduces a model agnostic naive approach to verify the context learnt and compared across three different language models on a context sensitive dataset.

As future objectives, we intend to extend the comparative study of language models for other NLP tasks like multi-label classification, text similarity and entity recognition etc. We look forward to extending the explainability work across datasets to visualize more context learning capabilities of language models. Also, other language models can be taken in account for comparative study.

## ACKNOWLEDGMENT

This research was performed in an internship program and was supported by Asset Management Group, Fidelity Investments. We thank colleagues from emerging tech team for their valuable suggestions that assisted the research.

## REFERENCES

[1] Kyunghyun Cho, Bart van Merrienboer, Caglar Gulcehre, Dzmitry Bahdanau, Fethi Bougares, Holger Schwenk, Yoshua Bengio, Learning Phrase Representations using RNN Encoder-Decoder for Statistical Machine Translation, 2014, eprint arXiv: 1406.1078

[2] Ashish Vaswani, Noam Shazeer, Niki Parmar, Jakob Uszkoreit, Llion Jones, Aidan N. Gomez, Lukasz Kaiser, Illia Polosukhin, Attention is all you need, June, 2017, preprint arXiv:1706.03762

[3] Jacob Devlin, Ming-Wei Chang, Kenton Lee, and Kristina Toutanova. 2018. BERT: Pre-training of Deep Bidirectional Transformers for Language Understanding. CoRR abs/1810.04805 (2018). arXiv:1810.04805

[4] Yinhan Liu, Myle Ott, Naman Goyal, Jingfei Du, Mandar Joshi, Danqi Chen, Omer Levy, Mike Lewis, Luke Zettlemoyer, Veselin Stoyanov, RoBERTa: A Robustly Optimized BERT Pretraining Approach, 2019, eprint arXiv: 1907.11692

[5] Victor Sanh and Lysandre Debut and Julien Chaumond and Thomas Wolf, DistilBERT, a distilled version of BERT: smaller, faster, cheaper and lighter, 2019, eprint arXiv: 1910.01108

[6] Zhenzhong Lan and Mingda Chen and Sebastian Goodman and Kevin Gimpel and Piyush Sharma and Radu Soricut, ALBERT: A Lite BERT for Self-supervised Learning of Language Representations, 2019, eprint arXiv: 1909.11942

[7] Kevin Clark and Minh-Thang Luong and Quoc V. Le and Christopher D. Manning, ELECTRA: Pre-training Text Encoders as Discriminators Rather Than Generators, 2020, eprint arXiv: 2003.10555

[8] Dogu Araci, FinBERT: Financial Sentiment Analysis with Pre-trained Language Models, 2019, eprint arXiv: 1908.10063

[9] Nabiha Asghar, Yelp Dataset Challenge: Review Rating Prediction, 2016, eprint arXiv: 1605.05362

[10] Alex Wang and Amanpreet Singh and Julian Michael and Felix Hill and Omer Levy and Samuel R. Bowman, GLUE: A Multi-Task Benchmark and Analysis Platform for Natural Language Understanding, 2018, eprint arXiv: 1804.07461

[11] Srikumar Krishnamoorthy, Sentiment Analysis of Financial News Articles using Performance Indicators, 2018, eprint arXiv: 1811.11008

[12] Jihyeok Kim, Reinald Kim Amplayo, Kyungjae Lee1 Sua Sung, Minji Seo, Seung-won Hwang, Categorical Metadata Representation for Customized Text Classification, Feb 2019, arXiv: 1902.05196

[13] Tang, Duyu and Qin, Bing and Liu, Ting, Learning Semantic Representations of Users and Products for Document Level Sentiment Classification, July 2015, https://www.aclweb.org/anthology/P15-1098

[14] Reinald Kim Amplayo, Rethinking Attribute Representation and Injection for Sentiment Classification, 2019, eprint arXiv: 1908.09590

[15] Iris Hendrickx, Su Nam Kim, Zornitsa Kozareva, Preslav Nakov, Diarmuid O Seaghdha, Sebastian Padok , Marco Pennacchiotti, Lorenza Romano, Stan Szpakowicz , SemEval-2010 Task 8: Multi-Way Classification of Semantic Relations Between Pairs of Nominals, 2010, S10-1006, Pages 28-33, Association for Computational Linguistics

[16] Livio Baldini Soares, Nicholas FitzGerald, Jeffrey Ling, Tom Kwiatkowski, Matching the Blanks:Distributional Similarity for Relation Learning, June 2019, arXiv: 1906.03158v1

[17] Cheng Li, Ye Tian, REDN Downstream Model Design re-trained Language Model for Relation Extraction Task, April 2020, arXiv: 2004.03786v1

[18] Mateusz Staniak, Przemysław Biecek, Explanations of model predictions with live and breakDown packages, April 2018, arXiv: 1804.01955

[19] Marco Tulio Ribeiro, Sameer Singh, Carlos Guestrin, "Why Should I Trust You?" Explaining the Predictions of Any Classifier, August, 2016 preprint arXiv:1602.04938

[20] Scott M. Lundberg, Su-In Lee, A Unified Approach to Interpreting Model Predictions, Nov, 2017, preprint arXiv: 1705.07874

[21] Avanti Shrikumar, Peyton Greenside, Anshul Kundaje, Learning Important Features Through Propagating Activation Differences, Oct, 2019 eprint arXiv: 1704.02685

[22] Kevin Clark, Urvashi Khandelwal, Omer Levy, Christopher D. Manning, What Does BERT Look At? An Analysis of BERT's Attention, June, 2019, preprint arXiv: 1705.07874

[23] Sarthak Jain, Byron C. Wallace, Attention is not Explanation, May, 2019, preprint arXiv: 1902.10186

[24] Sarah Wiegreffe, Yuval Pinter, Attention is not not Explanation, Sep, 2019, preprint arXiv: 1908.04626

[25] Tensorflow BERT Implementation Github Repository, [online], https://github.com/google-research/bert

[26] Tensorflow based ELECTRA Implementation Github Repository, [online], https://github.com/google-research/electra

[27] PyTorch DistilBERT, ALBERT, RoBERTa Implementation, [online], https://huggingface.co/transformers/

[28] Pekka Malo, Ankur Sinha, Pyry Talaka, Pekka Juhani Korhonen, Good Debt or Bad Debt: Detecting Semantic Orientations in Economic Texts, April 2014, 10.1002/asi.23062

[29] D. Greene and P. Cunningham. "Practical Solutions to the Problem of Diagonal Dominance in Kernel Document Clustering", Proc. ICML 2006

[30] Abhijeet Kumar, Rohit Gadia, Abhishek Pandey, Mridul Mishra, "Building Knowledge Graph using Pre-trained Language Model for Learning Entity-aware Relationships", GUCON IEEE Conference, 2020

# APPENDIX

| Getting Context – Comparison of Feature Importance across 3 Language Models (Top 3 words*) *words appear sequentially in order of significance (brighter denotes more importance) | | | |
|---|---|---|---|
| Sentences | BERT | RoBERTa | ELECTRA |
| The new factory working model and reorganizations would **decrease** Nokian Tyres' costs in the factory by EUR 30 million (USD 38.7 m). | costs by EUR | costs decrease in | costs decrease working |
| Pretax profit **decreased** by 37 % to EUR 193.1 mn from EUR 305.6 mn. | decreased from profit | decreased by EUR | decreased profit Pretax |
| Cash flow from operations totalled EUR 2.71 mn, compared to a **negative** EUR 0.83 mn in the corresponding period in 2008. | compared to EUR | compared negative EUR | compared negative Cash |
| In this case, the effect would be **negative** in Finland. | negative in Finland | negative in effect | effect negative be |
| The transaction doubles Tecnomens workforse and adds a fourth to their net **sales**. | doubles sales transaction | sales adds and | sales doubles fourth |
| Food **sales** totalled EUR 323.5 mn in October 2009, representing a decrease of 5.5 % from October 2008. | decrease 2008 sales | decrease 5.5 sales | decrease sales from |
| However, this **increases** signaling traffic which wastes network resources and allows fewer smartphones to connect. | wastes increases resources | wastes connect fewer | wastes increases which |
| The Department Store Division reported an **increase** in sales of 4 percent. | increase sales reported | increase reported an | increase sales Division |
| As a result, the Russia 's import restrictions on Finnish dairy companies will be **canceled** on 6 August 2010. | restrictions on will | canceled restrictions on | canceled restrictions result |
| Altogether Finnair has **canceled** over 500 flights because of the strike. | canceled flights strike | canceled because flights | canceled of because |
| In addition, the company will **reduce** a maximum of ten jobs. | jobs ten reduce | reduce jobs company | jobs reduce ten |
| UPM-Kymmene is working closely with its shipping agents to **reduce** fuel consumption and greenhouse gas emissions. | closely working reduce | reduce consumption UPM-Kymmene | reduce working greenhouse |
| Finnish insurance company Fennia and Kesko Group are ending their **loyal customer** cooperation. | ending cooperation loyal | ending customer loyal | ending customer Group |
| NTC has a geographical presence that complements Ramirent 's existing network and brings us closer to our **customers** in the East Bohemia region in the CzechRepublic. | closer brings customers | and complements brings | closer customers brings |
| The most **loyal customers** were found in the Bank of +land, with an index of 8.0. | loyal most index | loyal found of | loyal most 8.0 |
| The company initially estimated that it would **cut** up to 30 jobs. | jobs cut up | cut jobs to | cut jobs to |
| Thus the method will **cut** working costs, and will fasten the planning and building processes. | costs cut and | cut costs fasten | costs cut fasten |